\documentclass[sigconf]{acmart}
\usepackage[english]{babel}

\usepackage{amsmath}
\usepackage{graphicx}
\usepackage{tabularx}

\usepackage{graphicx}
\usepackage{xcolor}
\usepackage{xspace}
\usepackage{adjustbox}
\usepackage{csquotes}

\title{Zero-shot Entailment of Leaderboards for Empirical AI Research}

\author{Salomon Kabongo}
\orcid{0000-0002-0021-9729}
\affiliation{
  \institution{L3S Research Center, \\ Leibniz University of Hannover}
  \city{Hannover}
  \country{Germany}
}
\email{kabenamualu@l3s.de}

\author{Jennifer D'Souza}
\orcid{0000-0002-6616-9509}
\affiliation{%
  \institution{TIB Leibniz Information Centre for Science and Technology}
  \city{Hannover}
  \country{Germany}}
\email{jennifer.dsouza@tib.eu}

\author{S\"oren Auer}
\orcid{0000-0002-0698-2864}
\affiliation{%
  \institution{TIB Leibniz Information Centre for Science and Technology \& L3S Research Center}
  \city{Hannover}
  \country{Germany}}
\email{auer@tib.eu}

\begin{document}

\begin{abstract}
We present a large-scale empirical investigation of the zero-shot learning phenomena in a specific recognizing textual entailment (RTE) task category, i.e. the automated mining of \textsc{leaderboards} for Empirical AI Research. The prior reported state-of-the-art models for \textsc{leaderboards} extraction formulated as an RTE task, in a non-zero-shot setting, are promising with above 90\% reported performances. However, a central research question remains unexamined: \textit{did the models actually learn entailment?} Thus, for the experiments in this paper, two prior reported state-of-the-art models are tested out-of-the-box for their ability to \textit{generalize} or their capacity for \textit{entailment}, given \textsc{leaderboard} labels that were unseen during training. We hypothesize that if the models learned entailment, their zero-shot performances can be expected to be moderately high as well--perhaps, concretely, better than chance. As a result of this work, a zero-shot labeled dataset is created via distant labeling formulating the \textsc{leaderboard} extraction RTE task. 

\end{abstract}

\maketitle

\section{Introduction}


The recognizing textual entailment (RTE) task is defined as \enquote{a directional relation between two text fragments $T-H$, called text ($T$, the entailing text or context or premise), and hypothesis ($H$, the entailed text), so that a human being, with common understanding of language and common background knowledge, can infer that $H$ is most likely true on the basis of the content of $T$}~\cite{10.1007/11736790_9,10.1007/11736790_9,giampiccolo-etal-2007-third,giampiccolo2008fourth,bentivoglififth,Bentivogli2009TheSP,Bentivogli2011TheSP}. Proposed as an official community shared task for the first time in 2005~\cite{10.1007/11736790_9}, RTE has enjoyed a constantly growing popularity~\cite{chatzikyriakidis2017overview} in the NLP community, as it seems to work as a common framework in which to analyze, compare, and evaluate different techniques used in NLP applications to deal with semantic inference, a common issue shared by many NLP applications. In other words, different NLP applications such as reading comprehension, question answering, information extraction, machine translation, etc., can be recasted~\cite{glickman2006applied,white-etal-2017-inference,poliak-etal-2018-collecting} as an RTE task, thereby placing all tasks within a common evaluation paradigm.

In the realm of the RTE tasks collection, the task of automated mining of \textsc{leaderboards} for Empirical AI Research currently reports state-of-the-art (SOTA) performances above 90\% F1 as an RTE task~\cite{kabongo2021automated}. The extraction RTE task itself is formulated as follows. For \textsc{leaderboard} RTE, in each $T-H$ instance, the text $T$ corresponds to the scholarly article or a representation of the scholarly article from which the \textsc{leaderboard} should be extracted. The hypothesis $H$ corresponds to a (task, dataset, metric) tuple, TDM henceforward, obtained from a knowledge base (KB) of TDMs. In the context of the \textsc{leaderboard} extraction task, the extraction objective hypothesis $H$ has been variously formulated in prior work. E.g., either as (task, dataset, metric, method)~\cite{scirex} or (task, dataset, metric, score)~\cite{hou2019identification}. However, since the reported SOTA performance addresses the TDM triple objective, this is the focus of this work.

\begin{figure}
    \includegraphics[width=\linewidth, height=6cm]{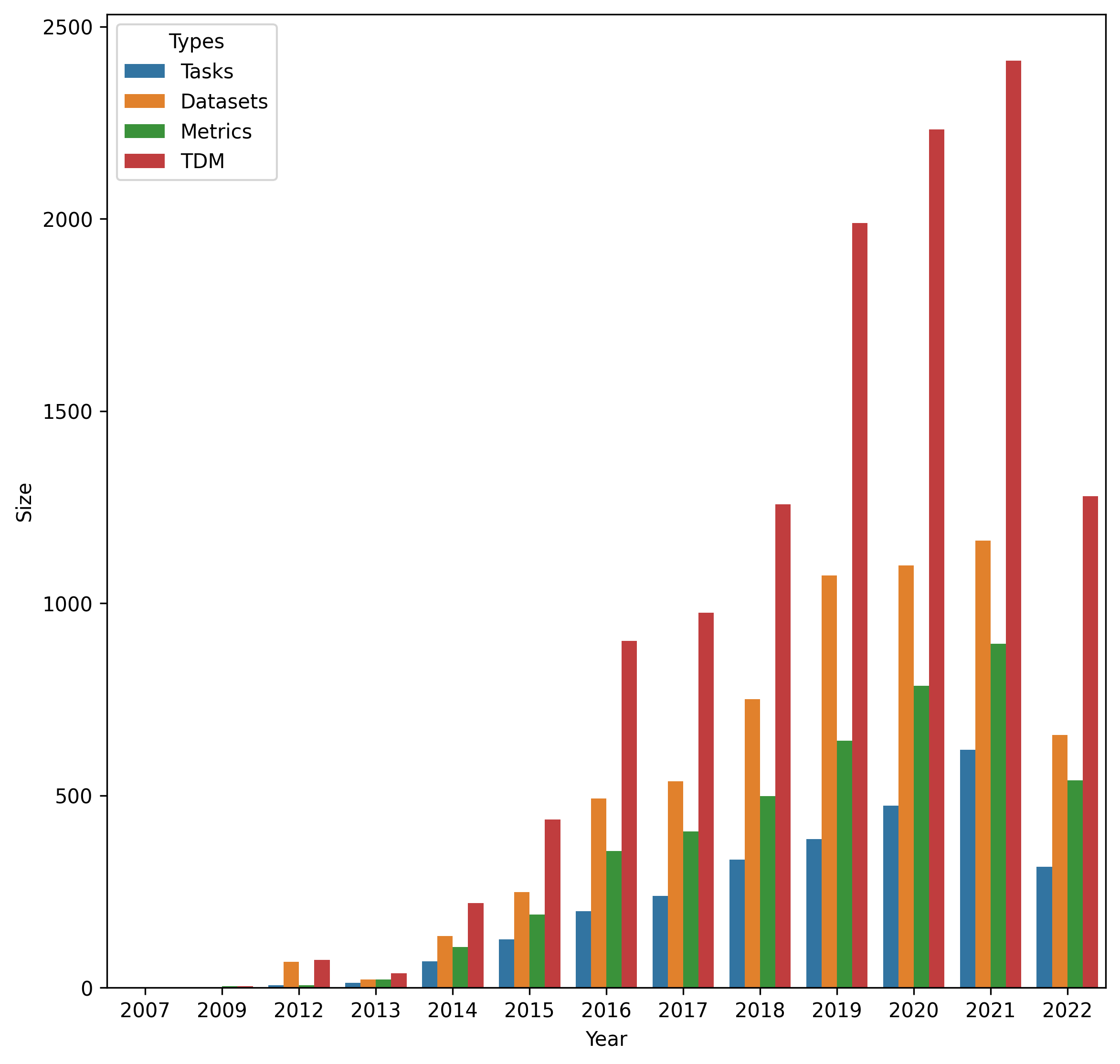}
    \caption{Rate of introduction of new tasks, datasets, metrics, and (task,dataset,metric) as tuples between 2007 and 2022.}\label{fig:a}
\end{figure}

With respect to the automated mining of \textsc{leaderboards} from scholarly articles task, there has been a recent surge in interest~\cite{hou2019identification,scirex,mondal2021end,hou2021tdmsci,citationie,kabongo2021automated}. In this era of the publications deluge where millions of articles are published each year~\cite{plume2011drowning}, it is no longer feasible for researchers to keep track of innovations by the traditional method of reading dozens of articles. Herein, \textsc{leaderboards} have proven a smart information access technology as easy progress trackers of models developed, for the empirical AI domain, in particular. \textsc{leaderboards} are often maintained in websites where models, their results, evaluation metrics, evaluation datasets, and code urls are reported and where specific research problems each have their own \textsc{leaderboard}. With the help of \textsc{leaderboards} now, researchers interested in developing new models on a problem have easy access to an overview on the current SOTA and performance trends of other models. Traditionally, they would need to scour through the discourse text of several papers and make mental notes of which models reported in which paper was better--a problem completely alleviated with \textsc{leaderboards}. Currently, \textsc{leaderboards} are mainly crowdsourced. E.g., \url{https://nlpprogress.com/}, \url{https://paperswithcode.com/}, or \url{https://orkg.org/benchmarks}. While the community curation of \textsc{leaderboards} have resulted in thousands of recorded results, note that it does not offer absolute TDM coverage guarantee. It can be that TDMs that are not noticed will be completely missed. As opposed to crowdsourcing, highly accurate automated extraction methods applied on article collections will not suffer from this limitation. Hence the vision is to complement the community curation of \textsc{leaderboards} with automated methods.

In lieu of this, a system~\cite{kabongo2021automated} reporting TDM extraction performances above 90\% would make it seem like a practical working solution is found. However, there remains a caveat in the model evaluation. The \textsc{leaderboard} extraction models have not been evaluated on unseen TDM labels. The RTE task, in the prior reported work~\cite{kabongo2021automated}, has only been evaluated on TDM labels that were seen during training. This leads to the formulation of the overarching research question (RQ) of this work: \textit{did the models actually learn entailment?} To obtain concrete insights for the RQ, the strategy we adopt is to test the two SOTA models out-of-the-box, namely ORKG-TDM$_{Bert}$ and ORKG-TDM$_{XLNet}$, in an exclusive zero-shot TDM extraction RTE setting. Our hypothesis is that if the models actually learnt an entailment task, they will be able to perform at least better than chance on novel unseen TDMs.  

Testing the models entailment ability is important for two main reasons. 1) The rapid growth rates of modern science is predicated on millions of scholarly articles published each year~\cite{bornmann2015growth}. This includes empirical investigations of new models released on different combinations of novel introduced tasks or datasets or metrics. To offer concrete insights into the novelty, \autoref{fig:a} illustrates the release rates of tasks, datasets, metrics, respectively, and TDM tuples per year. Thus automated extraction methods need to identify TDM tuples as they are introduced in the community without the need to go through the costly process of repeated retraining of the models as and when new TDM tuples are introduced. And, 2) consequently, one of the main advantages of training an RTE model is its ability to entail new information. Given that the SOTA \textsc{leaderboard} extractor~\cite{kabongo2021automated} is trained as an RTE model, if it successfully learnt an entailment task, it should be able to extract new TDMs. Thus in the face of the high number of novel TDMs introduced, only the TDMs KB would need to be expanded with new TDMs and the model would not need frequent retraining. To this end, this short paper is then a focused empirical analysis of two state-of-the-art \textsc{leaderboard} extractors called ORKG-TDM$_{Bert}$ and ORKG-TDM$_{XLNet}$ \cite{kabongo2021automated} that are originally trained as RTE models and, in this paper, are exclusively tested for their entailment abilities of new TDMs that were unseen during training. Thus, we investigate ORKG-TDM$_{Bert}$ and ORKG-TDM$_{XLNet}$ in a zero-shot RTE experimental setting.



\begin{table*}[!htb]
\begin{tabular}{|p{11cm}|p{4.1cm}|} \hline
 \textbf{Text ($T$)} & \textbf{Hypothesis ($H$)}\\\hline
{\stackbox[l]{{\color{red}3D Instances} as 1D Kernels ? We introduce a {\color{red}3D instance representation} … clustering algorithms in standard 3D instance segmentation pipelines.  Results show that DKNet outperforms the state of the arts on both ScanNetV2 and {\color{teal}S3DIS} datasets .. The others are evaluated via 6- fold cross validation. mRec mPre mWCov {\color{blue}mCov} Table 5.}} & \stackbox[l]{{\color{red}3D Instance Segmentation}; \\ {\color{teal}S3DIS}; \\ {\color{blue}mCov}}  \\ \hline
{\stackbox[l]{{\color{red}Oriented Object Detection in Aerial Images} with Box Boundary-Aware Vectors {\color{red}Oriented object detection in aerial images} is a challenging task … We evaluate our method on two public aerial image datasets: {\color{teal}DOTA} and HRSC2016 We use {\color{teal}DOTA-v1.0} dataset …  second best detection results in each column. 2 BC ST BD SV Plane Harbor LV TC RA SBF GTF Ship HC {\color{blue}mAP} SP Bridge 2 FPS Image Size AP Table 3}} & \stackbox[l]{{\color{red}Object Detection In Aerial Images}; \\ {\color{teal}DOTA}; \\ {\color{blue}mAP}}  \\ \hline
\end{tabular}


\caption{Recognizing textual entailment (RTE) instances as text-hypothesis $T-H$ pairs for the task of automated mining of \textsc{leaderboard}s. The Table shows one \textsc{leaderboard} annotation for the two papers ''3D Instances as 1D Kernels`` \citep{wu20223d} and ``Oriented Object Detection in Aerial Images with Box Boundary-Aware Vectors'' \citep{yi2021oriented}, respectively. \textsc{leaderboard} constitute the hypothesis ($H$) in the RTE task and are expressed as $({\color{red}task}; {\color{teal}dataset}; {\color{blue}metric})$ tuples. On the other hand, the text ($T$) shown in the first column are snippets of the DocTAET compressed paper representation comprising only the extracted paper title, abstract, experimental setup, and tabular information instead of the paper's full text.}
\label{table1}
\end{table*}


\section{Related Work}
\label{rel-work}


\textbf{Existing \textsc{leaderboard} extraction objectives.} The task is addressed via two different objectives. As 1) information extraction (IE) and 2) entailment (RTE). As an IE task, traditional pipelined based extractors have been implemented. For instance, the SciREX~\cite{scirex} framework, which aims to extract (dataset, metric, task, method) tuples, address it via five IE pipelined modules. They are: mention identification, salient mention identification, pairwise coreference resolution, mention clustering, and document-based 4-ary relation identification. Posited as document-level IE, it uses a 2-level BERT+BiLSTM method to get token representations which are passed to a CRF layer to identify mentions. In the next step, each mention is classified as being salient or not. Next, a coreference model is trained to cluster these mentions into entities. A final classification layer predicts relationships between 4-tuples of entities represented by their cluster embeddings. A closely related system, CitationIE~\cite{citationie}, extends the SciREX architecture by incorporating structural and textual citation information into the mention identification, salient entity classification, and and relation extraction modules. A highlight of these IE models is that the relations need not be provided in advance. Thus when trained on a sizeable proportion of data, these models should become capable of extracting relations between previously unseen entity tuples. But, based on their end-to-end pipelined extraction experimental results, however, these models report less than 25\% F1 which is contrary to expectations making the current IE-based models highly unsuitable for use in practice. 

On the other hand, works following the RTE objective depend on a predefined KB of TDM tuples as \textsc{leaderboards} used as candidate hypotheses ($H$) in $T-H$ pairs, instantiated as a paper text ($T$) and a TDM tuple ($H$) to determine one or more TDM hypotheses that can be entailed by a paper text. As the paper text representation $T$, the IBM-TDMS work~\cite{hou2019identification} defined a shorter representation feature instead of simply using the full-text of the article. As $T$ they only use a selection of the paper based on certain parts of the article such as abstract, table captions, experimental section, etc., where \textsc{leaderboards} or TDMs are most likely to be mentioned. They called this feature the DocTAET representation. Each paper was then classified with TDM tuple(s) or unknown, defined as an entailment between DocTAET ($T$) and the TDM ($H$) tuples. See \autoref{table1} for concrete examples of the data instances. Their model relied on the BERT$_{base}$ transformer encoder architecture~\cite{devlin2018bert} with a final classification layer for the entailed or not entailed decision. For a corpus of 332 papers, the TDM KB comprised 77 distinct tuples. They reported 67.8\% in F1--a significant improvement over the IE models performing at 25\% F1. The following work, called ORKG-TDM~\cite{kabongo2021automated}, built upon the work by IBM-TDMS. They used the same RTE task setup as well as the DocTAET paper text representation. However, they performed the task on a significantly larger dataset of 5,361 papers with a TDM KB of 1,850 unique instances. Further, they contrasted three distinct strategies of transformer models~\cite{vaswani2017attention}: BERT$_{base}$~\cite{devlin2018bert}, scientific BERT~\cite{beltagy2019scibert}, and XLNet~\cite{yang2019xlnet}. To date they report SOTA performance on the task with above 90\% in F1. 

Given the promise shown by RTE models over IE, by performing the SOTA RTE zero-shot evaluation in this work, we provide concrete evidence in the community on the robustness of the \textsc{leaderboard} extraction RTE models as actual entailment models.

\noindent{\textbf{Diverse NLP tasks recast as recognizing textual entailment (RTE).}} Entailment is seen as a core function implicit to various NLP applications to successfully perform a task. It is possible for various semantic classes of NLP tasks to be explicitly recast as an entailment task thus offering a common framework to analyze, compare, and evaluate different techniques. Since 2005, the task has been performed in annual challenges, called the Recognizing Textual Entailment (RTE) challenges, which have helped foster the interest of the research community in textual entailment. In these RTE-1 to 7 challenges \cite{10.1007/11736790_9,10.1007/11736790_9,giampiccolo-etal-2007-third,giampiccolo2008fourth,bentivoglififth,Bentivogli2009TheSP,Bentivogli2011TheSP}, common NLP applications like information retrieval (IR), comparable documents (CD), reading comprehension (RC), question answering (QA), information extraction (IE), machine translation (MT), and summarization (SUM) were recast as entailment tasks. The $T-H$ pairs were created by expert annotators in the different NLP application settings. E.g., considering an entailment instance for the QA task, $T$ was the text where the answer is found and $H$ is the question reformulated as an assertion with the answer in it. As another example, in the SUM task context, $T$ is a text description whose summary is the assertion hypothesis $H$. The RTE task itself then required the participating systems to decide, given two text snippets $T$ and $H$, whether $T$ entails $H$, where the $T-H$ pairs were recast equivalent and explicit entailment formulations of the different popular NLP applications. Three main aspects that informed the evolution of the tasks were: 1) RTE first postulated as two-way classification (entailment and unknown) was extended as a three-way classification task (entailment, contradiction and unknown). Further, 2) to situate RTE realistically, the length of the Ts were made longer in later task runs. In RTE-5, $T$s had lengths up to 100 words, whereas in RTE-4 the average length was about 40 words. This length was meant to represent the average portion of the source document that a reader would naturally select, such as a paragraph or a group of related sentences. Also, 3) later variants of the RTE task focused on search applications in the context of summarization, to retrieve all relevant texts $T$ from a collection of documents given hypothesis $H$ as queries. The development of search technologies in the later tasks further bolstered the interest of the research communities to investigating the recast RTE task formulations.

Inspired from the RTE challenges, there were other initiatives that recast common NLP tasks as RTE. To this end, the Diverse Natural Language Inference Collection (DNC)~\cite{white-etal-2017-inference,poliak-etal-2018-collecting} recast seven semantic phenomena from a total of 13 datasets into labeled RTE examples. The seven recast semantic phenomenon were the following: event factuality (EF), named entity recognition (NER), gendered anaphora resolution (GAR), lexicosyntactic inference (Lex), figurative language (Puns), relation extraction (RE), and subjectivity (Sentiment). Each of these seven tasks were reformulated in terms of the RTE $T-H$ tuples, where $T$ was the task sentence itself and the hypothesis $H$, informed by the underlying task, was a short, manually constructed sentence having an entailed or not-entailed resolution. E.g., for Lex, the premise is a sentence with various kinds of verb constructions about an event. Then each context sentence was paired with three hypotheses: a. That thing happened, b. That thing may or may not have happened, and c. That thing didn’t happen. And the entailment task was entailed or not for each of the $T-H$ a, b, and c pairs. As another example, for the Puns task, $T$ was the original sentence with some Name, as a template placeholder, of a person that either expressed or did not express a pun. The premise was then paired with two hypothesis: a. Name heard a pun, and b. Name did not hear a pun. Thus if the original sentence was labeled as containing a pun, the $T-H$ (a) pair is labeled as entailed and $T-H$ (b) pair is labeled as not entailed, otherwise the labels were swapped. 

Inspired by these various RTE task formulations, the \textsc{leaderboard} RTE task defined the hypothesis $H$ as the TDM tuple to be entailed from the text representation $T$ posited as the shorter and selective DocTAET feature of the original paper full text. Here, aligned with the later versions of the RTE challenges (RTE-5 onward), we are also testing the models in the realistic setting of longer $T$ from which to entail information. Offering a paragraph as $T$ as opposed to a single sentence has been shown to increase the RTE task complexity.


\begin{table*}[!t]
\begin{tabular}{lcccccc}
\hline
\textbf{}        & \textbf{Macro P} & \textbf{Macro R} & \textbf{Macro F1} & \textbf{Micro P} & \textbf{Micro R} & \textbf{Micro F1} \\ \hline
 {}                & \multicolumn{6}{c}{ORKG-TDM$_{Bert}$}                                                                  \\ \hline
{Fold-1} &  20.1 &	83.4 &	28.9 &	14.1 &	72.9 &	23.6 \\ \hline
{Fold-2} & 16.2 &	89 &	24.4 &	10.4 &	81.7 &	18.4     \\ \hline
{Average Fold 1 and Fold 2} &  18.2 &	86.2 &	26.7 &	12.3 &	77.3 &	21.0 \\ \hline

& \multicolumn{6}{c}{ORKG-TDM$_{XLNet}$} \\
\hline 
{Fold-1} & 14.3 &	86.6 &	21.9 &	9.2 &	78.1 &	16.5     \\ \hline
{Fold-2} & 14.9 &	86.4 &	22.7 &	10.1 &	76.8 &	17.8    \\ \hline
{Average Fold 1 and Fold 2} &  14.6 &	86.5 &	22.3 &	9.7 &	77.5 &	17.2  \\ \hline
\end{tabular}
\caption{Zero-shot \textsc{leaderboard} RTE results for the two best models, viz. ORKG-TDM$_{Bert}$ and ORKG-TDM$_{XLNet}$, from \citealt{kabongo2021automated} with reported performances above 90\% in F1 in non-zero-shot experimental evaluations.}
\label{tab:2 v2}
\end{table*}

\section{Zero-shot Experimental Setup}

\subsection{Our Zero-shot Corpus of \textsc{leaderboards}}
\label{corpus}

To test the reference trained models in a zero-shot setting, we first needed to create a suitable experimental corpus. \textit{The essential criteria to be satisfied by the new corpus was that none of the TDMs seen by the models of the reference work~\cite{kabongo2021automated} during training could be included in the new experimental corpus.} In other words, the experimental corpus for this paper's zero-shot labeling experiments needed to contain only unseen TDM labels for the ORKG-TDM$_{Bert}$ and ORKG-TDM$_{XLNet}$ trained models from the reference work~\cite{kabongo2021automated}. We addressed this by first downloading a new dump of community-curated papers of \textsc{leaderboards} (timestamp: Nov. 04, 2022) from \url{https://paperswithcode.com/}. Note we rely on the same data source, viz. \url{https://paperswithcode.com/}, as that of the reference work~\cite{kabongo2021automated}. The dump download of the reference work to train the models had a timestamp of May 10, 2021. By a simple difference of the latest dump with the earlier, we first obtained a corpus of papers that were newly introduced and that were not part of the reference work dataset. While this ensured unseen $T$s in the $T-H$ pairs, we also wanted to establish a corpus of unseen $H$s or TDMs to satisfy the zero-shot labeling criteria i.e. the entailment of new TDMs.

The next natural question is \textit{how did we define a zero-shot TDM label?} To arrive at this definition, we made the following critical observation. Had we defined a zero-shot TDM label as one where each of the task, dataset, and metric that constituted a TDM were all unseen in the training dataset of the reference work, this would have resulted in an empty zero-shot TDM labels set. It is a common phenomenon that the ratio of new metrics released every year is almost negligible. Further, the introduction of new datasets for existing tasks is far more frequent than the introduction of new tasks itself. This is reflected in the trends depicted in \autoref{fig:a}. Thus instead of adopting a strict criteria where all elements of TDM tuple were new, we instead defined a zero-shot TDM tuple as one where any of the task, dataset, or metric were unseen before. The new set of papers were further filtered to satisfy this criteria in their TDM annotations. To better clarify, a paper had to have only zero-shot TDM labels per our adopted definition to qualify as a candidate for the zero-shot corpus of \textsc{leaderboards}. While this resulted in a corpus of over 3000 papers, we randomly selected a subsample of 1000 papers which had a resulting set of 1925 zero-shot TDM labels including the label ``unknown.''

Note that this work is relegated purely to test evaluations, and therefore there is no training set. In other words, the sample of 1000 papers with 1925 zero-shot TDM labels constitute a testing corpus on which the ORKG-TDM$_{Bert}$ and ORKG-TDM$_{XLNet}$ trained models from the reference work~\cite{kabongo2021automated} are tested out-of-the-box.


\subsection{Two Test \textsc{leaderboard} RTE Models}

The two selected test models, viz. ORKG-TDM$_{Bert}$ and ORKG-TDM$_{XLNet}$ uses the standard transformer models~\cite{vaswani2017attention} fine-tuned for sequence pair classification, used here for RTE, with a [SEP] token between the the DocTAET paper representation as text $T$ and the TDM tuple as hypothesis $H$. The maximum input length is 512 for BERT~\cite{devlin2018bert} and 2000 for XLNET~\cite{yang2019xlnet}. 

\section{Results and Discussion}
\label{results}

Experimental results from the ORKG-TDM$_{Bert}$ and ORKG-TDM$_{XLNet}$ applied in a zero-shot RTE setting are shown in \autoref{tab:2 v2}. Since the original models were evaluated in a two-fold setting, in this work, we apply the models from their respective folds on our zero-shot corpus of \textsc{leaderboards}. This is how two-fold results are reported. Note the underlying test corpus in each fold is the same, only the trained models per fold differ. Our initial RQ was: \textit{did the models actually learn entailment?} We observe results contrary to our positive expectations based on the models strong performances reported in their non-zero-shot results. ORKG-TDM$_{Bert}$ in a non-zero-shot setting reported 90.8\% macro F1 and 91.8\% micro F1~\cite{kabongo2021automated}. The same model in a zero-shot setting reports performances significantly less than chance at 26.7\% macro F1 and 21\% micro F1. Likewise, the other SOTA model ORKG-TDM$_{XLNet}$ in a non-zero-shot setting reported 91.2\% macro F1 and 92.4\% micro F1. This model in the zero-shot setting reported 22.3\% macro F1 and 17.2\% micro F1. By observing the results in \autoref{tab:2 v2}, we see that the models have a very high recall at the cost of precision. Thus, for each $T$, the models entail most of the TDM hypothesis as true without being able to identify which TDM hypothesis the given $T$ actually entails.

We relegate the following two main reasons for the model low performances. 1) The RTE task is essentially a sequence-pair classification machine learning objective. In other words, whether the model learns a classification task or a true entailment task depends on the nature of the problem and cannot be enforced upon the model. Given the ORKG-TDM$_{Bert}$ and ORKG-TDM$_{XLNet}$ strong performances in the non-zero-shot setting versus poor performances in the zero-shot setting, we hypothesize that the reference models originally learned as a multi-class classification task and not an entailment task in its true sense. 2) The longer DocTAET $T$ representation further poses a challenge for the model to learn an entailment task. We hypothesize that a shorter one or two sentence representation would be a better approach toward realizing a true RTE model for \textsc{leaderboard} extraction.


\section{Conclusion}
\label{conclusion}

We have offered a comprehensive and realistic look at \textsc{leaderboard} extraction as an RTE task. We hope the results of this work can offer more informed decisions for future research on this theme. E.g., investigating transformer-model powered prompt-based methods instead that have shown promise as zero-shot learners~\cite{weifinetuned}. 


\bibliographystyle{ACM-Reference-Format}
\bibliography{refs}


\begin{thebibliography}{25}


\ifx \showCODEN    \undefined \def \showCODEN     #1{\unskip}     \fi
\ifx \showDOI      \undefined \def \showDOI       #1{#1}\fi
\ifx \showISBNx    \undefined \def \showISBNx     #1{\unskip}     \fi
\ifx \showISBNxiii \undefined \def \showISBNxiii  #1{\unskip}     \fi
\ifx \showISSN     \undefined \def \showISSN      #1{\unskip}     \fi
\ifx \showLCCN     \undefined \def \showLCCN      #1{\unskip}     \fi
\ifx \shownote     \undefined \def \shownote      #1{#1}          \fi
\ifx \showarticletitle \undefined \def \showarticletitle #1{#1}   \fi
\ifx \showURL      \undefined \def \showURL       {\relax}        \fi
\providecommand\bibfield[2]{#2}
\providecommand\bibinfo[2]{#2}
\providecommand\natexlab[1]{#1}
\providecommand\showeprint[2][]{arXiv:#2}

\bibitem[Beltagy et~al\mbox{.}(2019)]%
        {beltagy2019scibert}
\bibfield{author}{\bibinfo{person}{Iz Beltagy}, \bibinfo{person}{Kyle Lo},
  {and} \bibinfo{person}{Arman Cohan}.} \bibinfo{year}{2019}\natexlab{}.
\newblock \showarticletitle{SciBERT: A Pretrained Language Model for Scientific
  Text}. In \bibinfo{booktitle}{\emph{Proceedings of the 2019 Conference on
  Empirical Methods in Natural Language Processing and the 9th International
  Joint Conference on Natural Language Processing (EMNLP-IJCNLP)}}.
  \bibinfo{pages}{3615--3620}.
\newblock


\bibitem[Bentivogli et~al\mbox{.}(2009)]%
        {Bentivogli2009TheSP}
\bibfield{author}{\bibinfo{person}{Luisa Bentivogli}, \bibinfo{person}{Peter
  Clark}, \bibinfo{person}{Ido Dagan}, {and} \bibinfo{person}{Danilo
  Giampiccolo}.} \bibinfo{year}{2009}\natexlab{}.
\newblock \showarticletitle{The Sixth PASCAL Recognizing Textual Entailment
  Challenge}. In \bibinfo{booktitle}{\emph{Text Analysis Conference}}.
\newblock


\bibitem[Bentivogli et~al\mbox{.}(2011)]%
        {Bentivogli2011TheSP}
\bibfield{author}{\bibinfo{person}{Luisa Bentivogli}, \bibinfo{person}{Peter
  Clark}, \bibinfo{person}{Ido Dagan}, {and} \bibinfo{person}{Danilo
  Giampiccolo}.} \bibinfo{year}{2011}\natexlab{}.
\newblock \showarticletitle{The Seventh PASCAL Recognizing Textual Entailment
  Challenge}.
\newblock \bibinfo{journal}{\emph{Theory and Applications of Categories}}
  (\bibinfo{year}{2011}).
\newblock


\bibitem[Bentivogli et~al\mbox{.}({[n.\,d.]})]%
        {bentivoglififth}
\bibfield{author}{\bibinfo{person}{Luisa Bentivogli}, \bibinfo{person}{Ido
  Dagan}, \bibinfo{person}{Hoa~Trang Dang}, \bibinfo{person}{Danilo
  Giampiccolo}, {and} \bibinfo{person}{Bernardo Magnini}.}
  \bibinfo{year}{[n.\,d.]}\natexlab{}.
\newblock \showarticletitle{The Fifth PASCAL Recognizing Textual Entailment
  Challenge}.
\newblock  (\bibinfo{year}{[n.\,d.]}).
\newblock


\bibitem[Bornmann and Mutz(2015)]%
        {bornmann2015growth}
\bibfield{author}{\bibinfo{person}{Lutz Bornmann} {and}
  \bibinfo{person}{R{\"u}diger Mutz}.} \bibinfo{year}{2015}\natexlab{}.
\newblock \showarticletitle{Growth rates of modern science: A bibliometric
  analysis based on the number of publications and cited references}.
\newblock \bibinfo{journal}{\emph{Journal of the Association for Information
  Science and Technology}} \bibinfo{volume}{66}, \bibinfo{number}{11}
  (\bibinfo{year}{2015}), \bibinfo{pages}{2215--2222}.
\newblock


\bibitem[Chatzikyriakidis et~al\mbox{.}(2017)]%
        {chatzikyriakidis2017overview}
\bibfield{author}{\bibinfo{person}{Stergios Chatzikyriakidis},
  \bibinfo{person}{Robin Cooper}, \bibinfo{person}{Simon Dobnik}, {and}
  \bibinfo{person}{Staffan Larsson}.} \bibinfo{year}{2017}\natexlab{}.
\newblock \showarticletitle{An overview of Natural Language Inference Data
  Collection: The way forward?}. In \bibinfo{booktitle}{\emph{Proceedings of
  the Computing Natural Language Inference Workshop}}.
\newblock


\bibitem[Dagan et~al\mbox{.}(2005)]%
        {10.1007/11736790_9}
\bibfield{author}{\bibinfo{person}{Ido Dagan}, \bibinfo{person}{Oren Glickman},
  {and} \bibinfo{person}{Bernardo Magnini}.} \bibinfo{year}{2005}\natexlab{}.
\newblock \showarticletitle{The PASCAL Recognising Textual Entailment
  Challenge}. In \bibinfo{booktitle}{\emph{Proceedings of the First
  International Conference on Machine Learning Challenges: Evaluating
  Predictive Uncertainty Visual Object Classification, and Recognizing Textual
  Entailment}} (Southampton, UK) \emph{(\bibinfo{series}{MLCW'05})}.
  \bibinfo{publisher}{Springer-Verlag}, \bibinfo{address}{Berlin, Heidelberg},
  \bibinfo{pages}{177–190}.
\newblock
\showISBNx{3540334270}
\urldef\tempurl%
\url{https://doi.org/10.1007/11736790_9}
\showDOI{\tempurl}


\bibitem[Giampiccolo et~al\mbox{.}(2008)]%
        {giampiccolo2008fourth}
\bibfield{author}{\bibinfo{person}{Danilo Giampiccolo},
  \bibinfo{person}{Hoa~Trang Dang}, \bibinfo{person}{Bernardo Magnini},
  \bibinfo{person}{Ido Dagan}, \bibinfo{person}{Elena Cabrio}, {and}
  \bibinfo{person}{Bill Dolan}.} \bibinfo{year}{2008}\natexlab{}.
\newblock \showarticletitle{The Fourth PASCAL Recognizing Textual Entailment
  Challenge.}. In \bibinfo{booktitle}{\emph{TAC}}.
\newblock


\bibitem[Giampiccolo et~al\mbox{.}(2007)]%
        {giampiccolo-etal-2007-third}
\bibfield{author}{\bibinfo{person}{Danilo Giampiccolo},
  \bibinfo{person}{Bernardo Magnini}, \bibinfo{person}{Ido Dagan}, {and}
  \bibinfo{person}{Bill Dolan}.} \bibinfo{year}{2007}\natexlab{}.
\newblock \showarticletitle{The Third {PASCAL} Recognizing Textual Entailment
  Challenge}. In \bibinfo{booktitle}{\emph{Proceedings of the {ACL}-{PASCAL}
  Workshop on Textual Entailment and Paraphrasing}}.
  \bibinfo{publisher}{Association for Computational Linguistics},
  \bibinfo{address}{Prague}, \bibinfo{pages}{1--9}.
\newblock
\urldef\tempurl%
\url{https://aclanthology.org/W07-1401}
\showURL{%
\tempurl}


\bibitem[Glickman(2006)]%
        {glickman2006applied}
\bibfield{author}{\bibinfo{person}{Oren Glickman}.}
  \bibinfo{year}{2006}\natexlab{}.
\newblock \bibinfo{booktitle}{\emph{Applied textual entailment}}.
\newblock \bibinfo{publisher}{Citeseer}.
\newblock


\bibitem[Hou et~al\mbox{.}(2019)]%
        {hou2019identification}
\bibfield{author}{\bibinfo{person}{Yufang Hou}, \bibinfo{person}{Charles
  Jochim}, \bibinfo{person}{Martin Gleize}, \bibinfo{person}{Francesca Bonin},
  {and} \bibinfo{person}{Debasis Ganguly}.} \bibinfo{year}{2019}\natexlab{}.
\newblock \showarticletitle{Identification of Tasks, Datasets, Evaluation
  Metrics, and Numeric Scores for Scientific Leaderboards Construction}. In
  \bibinfo{booktitle}{\emph{Proceedings of the 57th Annual Meeting of the
  Association for Computational Linguistics}}. \bibinfo{publisher}{Association
  for Computational Linguistics}, \bibinfo{address}{Florence, Italy},
  \bibinfo{pages}{5203--5213}.
\newblock
\urldef\tempurl%
\url{https://doi.org/10.18653/v1/P19-1513}
\showDOI{\tempurl}


\bibitem[Hou et~al\mbox{.}(2021)]%
        {hou2021tdmsci}
\bibfield{author}{\bibinfo{person}{Yufang Hou}, \bibinfo{person}{Charles
  Jochim}, \bibinfo{person}{Martin Gleize}, \bibinfo{person}{Francesca Bonin},
  {and} \bibinfo{person}{Debasis Ganguly}.} \bibinfo{year}{2021}\natexlab{}.
\newblock \showarticletitle{TDMSci: A Specialized Corpus for Scientific
  Literature Entity Tagging of Tasks Datasets and Metrics}. In
  \bibinfo{booktitle}{\emph{Proceedings of the 16th Conference of the European
  Chapter of the Association for Computational Linguistics: Main Volume}}.
  \bibinfo{pages}{707--714}.
\newblock


\bibitem[Jain et~al\mbox{.}(2020)]%
        {scirex}
\bibfield{author}{\bibinfo{person}{Sarthak Jain}, \bibinfo{person}{Madeleine
  van Zuylen}, \bibinfo{person}{Hannaneh Hajishirzi}, {and} \bibinfo{person}{Iz
  Beltagy}.} \bibinfo{year}{2020}\natexlab{}.
\newblock \showarticletitle{SciREX: A Challenge Dataset for Document-Level
  Information Extraction}. In \bibinfo{booktitle}{\emph{Proceedings of the 58th
  Annual Meeting of the Association for Computational Linguistics}}.
  \bibinfo{pages}{7506--7516}.
\newblock


\bibitem[Kabongo et~al\mbox{.}(2021)]%
        {kabongo2021automated}
\bibfield{author}{\bibinfo{person}{Salomon Kabongo}, \bibinfo{person}{Jennifer
  D’Souza}, {and} \bibinfo{person}{S{\"o}ren Auer}.}
  \bibinfo{year}{2021}\natexlab{}.
\newblock \showarticletitle{Automated mining of leaderboards for empirical ai
  research}. In \bibinfo{booktitle}{\emph{International Conference on Asian
  Digital Libraries}}. Springer, \bibinfo{pages}{453--470}.
\newblock


\bibitem[Kenton and Toutanova(2019)]%
        {devlin2018bert}
\bibfield{author}{\bibinfo{person}{Jacob Devlin Ming-Wei~Chang Kenton} {and}
  \bibinfo{person}{Lee~Kristina Toutanova}.} \bibinfo{year}{2019}\natexlab{}.
\newblock \showarticletitle{Bert: Pre-training of deep bidirectional
  transformers for language understanding}. In
  \bibinfo{booktitle}{\emph{Proceedings of naacL-HLT}}.
  \bibinfo{pages}{4171--4186}.
\newblock


\bibitem[Mondal et~al\mbox{.}(2021)]%
        {mondal2021end}
\bibfield{author}{\bibinfo{person}{Ishani Mondal}, \bibinfo{person}{Yufang
  Hou}, {and} \bibinfo{person}{Charles Jochim}.}
  \bibinfo{year}{2021}\natexlab{}.
\newblock \showarticletitle{End-to-end construction of NLP knowledge graph}. In
  \bibinfo{booktitle}{\emph{Findings of the Association for Computational
  Linguistics: ACL-IJCNLP 2021}}. \bibinfo{pages}{1885--1895}.
\newblock


\bibitem[Plume(2011)]%
        {plume2011drowning}
\bibfield{author}{\bibinfo{person}{Andrew Plume}.}
  \bibinfo{year}{2011}\natexlab{}.
\newblock \showarticletitle{Drowning in the publication deluge?}
\newblock \bibinfo{journal}{\emph{Research Trends}} \bibinfo{volume}{1},
  \bibinfo{number}{22} (\bibinfo{year}{2011}), \bibinfo{pages}{7}.
\newblock


\bibitem[Poliak et~al\mbox{.}(2018)]%
        {poliak-etal-2018-collecting}
\bibfield{author}{\bibinfo{person}{Adam Poliak}, \bibinfo{person}{Aparajita
  Haldar}, \bibinfo{person}{Rachel Rudinger}, \bibinfo{person}{J.~Edward Hu},
  \bibinfo{person}{Ellie Pavlick}, \bibinfo{person}{Aaron~Steven White}, {and}
  \bibinfo{person}{Benjamin Van~Durme}.} \bibinfo{year}{2018}\natexlab{}.
\newblock \showarticletitle{Collecting Diverse Natural Language Inference
  Problems for Sentence Representation Evaluation}. In
  \bibinfo{booktitle}{\emph{Proceedings of the 2018 Conference on Empirical
  Methods in Natural Language Processing}}. \bibinfo{publisher}{Association for
  Computational Linguistics}, \bibinfo{address}{Brussels, Belgium},
  \bibinfo{pages}{67--81}.
\newblock
\urldef\tempurl%
\url{https://doi.org/10.18653/v1/D18-1007}
\showDOI{\tempurl}


\bibitem[Vaswani et~al\mbox{.}(2017)]%
        {vaswani2017attention}
\bibfield{author}{\bibinfo{person}{Ashish Vaswani}, \bibinfo{person}{Noam
  Shazeer}, \bibinfo{person}{Niki Parmar}, \bibinfo{person}{Jakob Uszkoreit},
  \bibinfo{person}{Llion Jones}, \bibinfo{person}{Aidan~N Gomez},
  \bibinfo{person}{{\L}ukasz Kaiser}, {and} \bibinfo{person}{Illia
  Polosukhin}.} \bibinfo{year}{2017}\natexlab{}.
\newblock \showarticletitle{Attention is all you need}. In
  \bibinfo{booktitle}{\emph{Advances in neural information processing
  systems}}. \bibinfo{pages}{5998--6008}.
\newblock


\bibitem[Viswanathan et~al\mbox{.}(2021)]%
        {citationie}
\bibfield{author}{\bibinfo{person}{Vijay Viswanathan}, \bibinfo{person}{Graham
  Neubig}, {and} \bibinfo{person}{Pengfei Liu}.}
  \bibinfo{year}{2021}\natexlab{}.
\newblock \showarticletitle{{C}itation{IE}: Leveraging the Citation Graph for
  Scientific Information Extraction}. In \bibinfo{booktitle}{\emph{Proceedings
  of the 59th Annual Meeting of the Association for Computational Linguistics
  and the 11th International Joint Conference on Natural Language Processing
  (Volume 1: Long Papers)}}. \bibinfo{publisher}{Association for Computational
  Linguistics}, \bibinfo{address}{Online}, \bibinfo{pages}{719--731}.
\newblock
\urldef\tempurl%
\url{https://doi.org/10.18653/v1/2021.acl-long.59}
\showDOI{\tempurl}


\bibitem[Wei et~al\mbox{.}({[n.\,d.]})]%
        {weifinetuned}
\bibfield{author}{\bibinfo{person}{Jason Wei}, \bibinfo{person}{Maarten Bosma},
  \bibinfo{person}{Vincent Zhao}, \bibinfo{person}{Kelvin Guu},
  \bibinfo{person}{Adams~Wei Yu}, \bibinfo{person}{Brian Lester},
  \bibinfo{person}{Nan Du}, \bibinfo{person}{Andrew~M Dai}, {and}
  \bibinfo{person}{Quoc~V Le}.} \bibinfo{year}{[n.\,d.]}\natexlab{}.
\newblock \showarticletitle{Finetuned Language Models are Zero-Shot Learners}.
  In \bibinfo{booktitle}{\emph{International Conference on Learning
  Representations}}.
\newblock


\bibitem[White et~al\mbox{.}(2017)]%
        {white-etal-2017-inference}
\bibfield{author}{\bibinfo{person}{Aaron~Steven White},
  \bibinfo{person}{Pushpendre Rastogi}, \bibinfo{person}{Kevin Duh}, {and}
  \bibinfo{person}{Benjamin Van~Durme}.} \bibinfo{year}{2017}\natexlab{}.
\newblock \showarticletitle{Inference is Everything: Recasting Semantic
  Resources into a Unified Evaluation Framework}. In
  \bibinfo{booktitle}{\emph{Proceedings of the Eighth International Joint
  Conference on Natural Language Processing (Volume 1: Long Papers)}}.
  \bibinfo{publisher}{Asian Federation of Natural Language Processing},
  \bibinfo{address}{Taipei, Taiwan}, \bibinfo{pages}{996--1005}.
\newblock
\urldef\tempurl%
\url{https://aclanthology.org/I17-1100}
\showURL{%
\tempurl}


\bibitem[Wu et~al\mbox{.}(2022)]%
        {wu20223d}
\bibfield{author}{\bibinfo{person}{Yizheng Wu}, \bibinfo{person}{Min Shi},
  \bibinfo{person}{Shuaiyuan Du}, \bibinfo{person}{Hao Lu},
  \bibinfo{person}{Zhiguo Cao}, {and} \bibinfo{person}{Weicai Zhong}.}
  \bibinfo{year}{2022}\natexlab{}.
\newblock \showarticletitle{3D Instances as 1D Kernels}. In
  \bibinfo{booktitle}{\emph{European Conference on Computer Vision}}. Springer,
  \bibinfo{pages}{235--252}.
\newblock


\bibitem[Yang et~al\mbox{.}(2019)]%
        {yang2019xlnet}
\bibfield{author}{\bibinfo{person}{Zhilin Yang}, \bibinfo{person}{Zihang Dai},
  \bibinfo{person}{Yiming Yang}, \bibinfo{person}{Jaime Carbonell},
  \bibinfo{person}{Russ~R Salakhutdinov}, {and} \bibinfo{person}{Quoc~V Le}.}
  \bibinfo{year}{2019}\natexlab{}.
\newblock \showarticletitle{Xlnet: Generalized autoregressive pretraining for
  language understanding}.
\newblock \bibinfo{journal}{\emph{Advances in neural information processing
  systems}}  \bibinfo{volume}{32} (\bibinfo{year}{2019}).
\newblock


\bibitem[Yi et~al\mbox{.}(2021)]%
        {yi2021oriented}
\bibfield{author}{\bibinfo{person}{Jingru Yi}, \bibinfo{person}{Pengxiang Wu},
  \bibinfo{person}{Bo Liu}, \bibinfo{person}{Qiaoying Huang},
  \bibinfo{person}{Hui Qu}, {and} \bibinfo{person}{Dimitris Metaxas}.}
  \bibinfo{year}{2021}\natexlab{}.
\newblock \showarticletitle{Oriented object detection in aerial images with box
  boundary-aware vectors}. In \bibinfo{booktitle}{\emph{Proceedings of the
  IEEE/CVF Winter Conference on Applications of Computer Vision}}.
  \bibinfo{pages}{2150--2159}.
\newblock


\end{thebibliography}

\end{document}